\documentclass[sigconf]{acmart}

\usepackage{booktabs} 
\usepackage{graphicx}
\usepackage{amsmath,amssymb} 
\usepackage{epsfig}
\usepackage{subfigure}
\usepackage{epstopdf}
\usepackage{multirow}
\usepackage{array}

\setcopyright{rightsretained}

\begin{document}
\title{Online action recognition based on incremental learning of weighted covariance descriptors}

%
\author{Chang Tang}
\affiliation{%
  \institution{School of Computer Science, \\China University of Geosciences}
  \city{Wuhan} 
  \state{Hubei, China} 
  \postcode{430074}
}
\email{happytangchang@gmail.com}

\author{Pichao Wang}
\authornote{Corresponding author.}
\affiliation{%
  \institution{Advanced Multimedia Research Lab, University of Wollongong}
  \city{Wollongong} 
  \state{NSW, Australia} 
  \postcode{2500}
}
\email{pw212@uowmail.edu.au}

\author{Wanqing Li}
\affiliation{%
  \institution{Advanced Multimedia Research Lab, University of Wollongong}
  \city{Wollongong} 
  \state{NSW, Australia}
  \postcode{2500}
}
\email{wanqing@uow.edu.au}

\begin{abstract}
Different from traditional action recognition based on video segments, online action recognition aims to recognize actions from unsegmented streams of data in a continuous manner.  One way for online recognition is based on the evidence accumulation over time to make predictions from stream videos. This paper presents a fast yet effective method to recognize actions from stream of noisy skeleton data, and a novel weighted covariance descriptor is adopted to accumulate evidence. In particular, a fast incremental updating method for the weighted covariance descriptor is developed for accumulation of temporal information and online prediction. The weighted covariance descriptor takes the following principles into consideration:  past frames have less contribution for recognition and recent and informative frames such as key frames contribute more to the recognition. The online recognition is achieved using a simple nearest neighbor search against a set of offline trained action models. Experimental results on MSC-12 Kinect Gesture dataset and our newly constructed online action recognition dataset have demonstrated the efficacy of the proposed method.
\end{abstract}

%
%



\maketitle

\section{Introduction}

Human action recognition is an active research topic in computer vision due to its wide range of potential applications, viz. surveillance, video games, video indexing and search, and human-robot interaction. In the last decade many approaches have been proposed to recognize actions from monocular or RGB video sequences~\cite{aggarwal2011human}. However, these methods face the difficulties posed by changes in illumination, variations in viewpoint, occlusion and cluttered background. Perhaps more importantly, these methods are somewhat impaired by the loss of 3D information in conventional video.

Since the release of low-cost RGB-D sensors such as Microsoft Kinect~\texttrademark sensors, many efforts and advances have been made on action recognition from depth maps. Compared with RGB data, depth maps have several advantages for action recognition, typically, being insensitive to illumination changes and reliable to estimate body silhouette and skeleton~\cite{shotton2013real}.


Many methods have been proposed for recognizing actions from depth or skeleton data~\cite{aggarwal2014human,li2010action}. These methods are often based on hand-crafted features, such as depth-map-based~\cite{yang2012recognizing}, skeleton joints~\cite{vemulapalli2014human,zanfir2013moving} or body parts~\cite{shotton2013real}, cloud points~\cite{wang2012robust,rahmani2014hopc}, local interest points~\cite{xia2013spatio},and surface normals~\cite{oreifej2013hon4d,yang2014super}. With the development of deep learning approach, several works have been proposed based on Convolutional Neural Networks (CNNs)~\cite{wang2015convnets,wang2016mm,wang2016action} and Recurrent Neural Networks (RNNs)~\cite{donahue2015long,du2015hierarchical,shahroudy2016ntu}. However, all of these methods only focus on classifying actions from segmented sequences of input data, with each segment corresponding to one single action. They assumed that all the instances, training or testing, are temporally segmented before recognition. This assumption is usually not valid when data are streamed in real-time and recognition has to be conducted online, where boundaries between different kinds of actions within the stream are unknown. 

Generally speaking, there are two main approaches to online recognition based on the types of video representation: frame-level representation and sliding window based representation. In the former, the stream video is represented by a series of frame-level descriptors where each descriptor is extracted from one or several frames in a short temporal interval~\cite{yu2014discriminative,jiang2014online,miranda2014online,hu2016discriminative}. The advantage of this representation is its simplicity, without detecting the start and end frame of each action and online recognition is achieved by aggregating frame-based classification. However, this representation tends to neglect the temporal coherence, which is vital for recognizing complex actions, and frame-based classification is prone to error. 

Sliding window based methods~\cite{kviatkovsky2014online,kulkarni2014continuous,zhu2016online} are a simple extension of segmented-based action recognition methods. They often consider the temporal coherence within the window for prediction and the window-based predictions are further fused to achieve online recognition. However, the performance of these methods are sensitive to the window size which depends on actions and is hard to set. Either too large or too small window size could lead to a significant drop in recognition. In addition, in previous sliding window based methods, all the frames in the sliding window are often considered equally important which are not justifiable. When the window is large and covers more than one actions, the past frames should contribute less to the recognition of recent frames. Moreover, frames that are discriminative should also contribute more to the recognition than non-discriminative frames. Motivated by the mathematical properties of covariance descriptors~\cite{hussein2013human,kviatkovsky2014online,sanin2013spatio} and its being able to incrementally updated, 
this paper proposes a fast online action recognition method from skeleton data based on weighted covariance descriptors. The method assumes that segmented and labeled action instances are available for offline training and recognition is to be performed in an online manner. 
To facilitate the online recognition, an incremental learning of the weighted covariance descriptors is developed by taking into consideration the importance of frames with respect to their temporal order and discrimination. Such an incremental learning provides an effective mechanism to accumulate information over time for recognition.
Experimental results on MSRC-12 Kinect Gesture dataset~\cite{fothergill2012instructing} and on our newly collected online action recognition dataset have demonstrated the efficacy of the proposed method. The new dataset will be released to the public upon the acceptance of this paper.

The contributions of this paper are summarized as follows: 1) An effective evidence accumulation based fast online action recognition method is proposed based on weighted covariance descriptors, which is the first attempt to adopt weighted covariance descriptor for stream based action recognition; 2) a fast incremental learning of the covariance descriptors with two kinds of weights capturing both temporal order, i.e. past frames are gradually ``forgotten", and the discrimination of frames; 3) a new RGB-D online action recognition dataset was created and will be released to the public; 4) state-of-the-art results are achieved by the proposed method on the two datasets. 

The remainder of this paper is organized  as follows. Section~\ref{related} presents the related work. Section~\ref{method} describes the proposed method. Experimental results on the two datasets are presented in Section~\ref{experimental}. Section~\ref{conclusion} concludes the paper with discussion on the future work.

\section{Related work}
\label{related}
Recently, several online action recognition methods have been proposed. These methods can be mainly categorized into to two classes: frame-level based  and sliding window based. 

A typical frame-based method was introduced by Miranda et al.~\cite{miranda2014online} for real-time gesture recognition from noisy skeleton streams. In their method, key poses were identified by the descriptors composed of angular representation of the skeleton joints, and the gesture was labeled on-the-fly from the key pose sequence with a decision forest; Zhao et al.~\cite{zhao2014structured} proposed a new feature, structured streaming skeleton, for online human gesture recognition, which is constructed for each frame through dynamic matching; De Rosa et al.~\cite{DeRosa2014} proposed a novel algorithm for online nonparametric recognition built upon a recent nonparametric regression method. Their online action recognition system can be combined with any set of frame-by-frame feature descriptors; Yu et al.~\cite{yu2014discriminative} proposed a discriminative orderlet mining method for real-time recognition of human-object interaction by using both depth maps and skeleton joints; Hu et al.~\cite{hu2016discriminative} represented an approach for online human action recognition, where the videos were represented by frame-level descriptors. They proposed a method to discover an action states from frame-level descriptors. However, these frame based methods tend to neglect the temporal coherence, which is vital for recognizing complex actions. 

Sliding window based methods, are a simple extension of segmented based action recognition in which the temporal coherence inside the window is well taken into consideration. For example, in~\cite{hoai2011joint}, action segmentation and recognition were jointly performed based on a discriminative temporal extension of the spatial bag-of-words model; Minhas et al.~\cite{minhas2012incremental} approximated the shape of a human by adaptively changing intensity histograms to extract pyramid histograms of oriented gradient features. They then examined incremental learning as an overlooked obstruction to the implementation of reliable real-time recognition; Vieira et al.~\cite{vieira2014improvement} constructed a new high dimensional feature vector, called Space-Time Occupancy Pattern (STOP) by dividing space and time axes into multiple segments. Online action recognition was performed by combining depth maps with skeletons; Kulkarni et al.~\cite{kulkarni2014continuous} built on the well known dynamic time warping (DTW) framework and devised a visual alignment technique, namely dynamic frame warping (DFW), which performed isolated recognition based on per-frame representation of videos; Hasan et al.~\cite{hasan2014continuous} proposed a continuous human activity learning framework from streaming videos by intricately tying together deep networks and active learning. In their work, given the segmented activities from streaming videos, they learned features in an unsupervised manner using deep networks and use active learning to reduce the amount of manual labeling of classes; In~\cite{zhu2016online}, the authors did not detect the start and end points of each human action explicitly, but segmented feature sequences online, and employed a variable-length MEMM method to recognize human actions based on the online model matching results of feature segments; Bloom et al.~\cite{bloom2016hierarchical} proposed a hierarchical transfer learning algorithm for online detection of compound actions for robust action recognition. Transfer learning was employed to allow the tasks of action segmentation and modelling, and model adaptation was used to improve performance on complex datasets.

It is worth noting that in online action recognition past frames from a stream may not necessarily contribute to the classification of current frames. This is especially true when the past frames are actually from an action different from the current action. In addition, frames in the period of a same action would have different discriminative power to the classification of the frames. Therefore, a mechanism is needed to address these two factors for online action recognition.  

Covariance descriptor which was originally designed for object detection, has been successfully used for 2D and 3D action recognition~\cite{harandi2012kernel,hussein2013human,kviatkovsky2014online,sanin2013spatio}. In the work of Harandi et al.~\cite{harandi2012kernel}, the covariance descriptors were mapped into the Euclidean space by a Riemannian locality preserving projection (RLPP) technique and action classification was performed by standard classification methods. Sanin et al.~\cite{sanin2013spatio} used the RLPP with AdaBoost to learn a set of covariance descriptors and obtained impressive results in the recognition of a specific set of actions. Hussein et al.~\cite{hussein2013human} used a temporal hierarchy of covariance matrices on 3D joint locations over time as a discriminative descriptor for a sequence. They obtained good action recognition results by using linear SVM on the descriptors. Kviatkovsky et al.~\cite{kviatkovsky2014online} introduced an incremental updating rule for covariance matrices and used a sliding window to construct a covariance feature descriptor, the final frame by frame online action recognition was performed through nearest neighbor classification. This paper also uses the incremental covariance update rule but our work is different from what Kviatkovsky et al.~\cite{kviatkovsky2014online} reported. In their work, action class of a current frame was detected based on the covariance matrix constructed from previous $W-1$ frames and current frame; $W$ is the sliding window size. So if the current frame is the beginning of a new action, it is usually wrongly classified to the previous action. This situation arises because most of the information captured by the current covariance matrix is from previous action frames.

Another issue with their work is that they considered each frame in the sliding window equally important. Commonly,  current frames are representative for actions being performed, hence, they are more important, than past frames. Then key-frame based action recognition has demonstrated that frames within an instance of an action are not equally discriminative, for instance, frames of neutral poses often do not contribute to the classification.

This paper addresses the two drawbacks discussed above by proposing a weighted covariance descriptor and its incremental updating that take into consideration both the temporal order and discrimination of frames. 

\section{The Proposed Method}\label{method}
\label{Framework}
Suppose there are $L$ possible action classes and $M$ segmented training action instances, and each training action instance corresponds to one of the $L$ action class. The action-label set can be denoted by $\mathcal{L} = \{ l\} _{l = 1}^L$. Let $\left\{ {M_n^l} \right\}_{n = 1}^{{N_l}}$ denote the set
of $N_l$ single-action instances of class $l$, where ${N_1} +  \cdots  + {N_L} = M$. Given a test video sequence $V$ with an unknown order and number
of actions $I$, its unknown label sequence is represented as $Z = ({z_1}, \cdots ,{z_i}, \cdots ,{z_I})$, $z_i \in \mathcal{L}$; the boundaries between two consecutive actions are also unknown. In practice, we cannot access the whole test video sequence at once; only one frame of the test streaming video is available at time $t$. The task of online action recognition is to decide what action class the human is performing at any time $t$ using sufficient information of the previous frames before $t$, and find a partition between two consecutive different actions. If sufficient information is accumulated at time $t$ to make a decision, we determine that the subject is performing an action $l$, otherwise continue accumulating information from time $t + 1$.

\subsection{The Covariance Descriptor}
\label{CovarianceDescriptor}
Let $\mathbf{S}=[{s_1},{s_2}, \cdots ,{s_n}]$ be the data matrix composed of $n$ feature vectors, each feature vector containing $d$ feature variables. The data matrix can be represented by the $d\times d$ sample covariance matrix $C$ as follows:
\begin{equation}\label{CovDef}
C = \frac{1}{{n - 1}}\sum\limits_{i = 1}^n {\left( {{s_i} - \bar s} \right)} {\left( {{s_i} - \bar s} \right)^T},
\end{equation}
where $\bar s$ is the mean of the feature vectors. The diagonal entries of the covariance matrix represent the variance of each individual feature variable, while the non-diagonal entries are their respective correlations. Our motivation of using covariance as a feature descriptor can be
summarized as follows. Firstly, the covariance matrix can be calculated in an incremental manner so as to accumulate evidence over time and it is simple and fast to compute, thus meeting the essential requirement of online recognition. Secondly, the covariance matrix captures information about the shape and distribution of the set of feature variables. The feature variables of different actions generally have different distributions, so covariance matrices can discriminate different actions.

\subsection{Overview of the method}
\label{Overview}
In this section, the details of the proposed method are presented. There are two main phases in the proposed method: offline training and online recognition.

\subsubsection{Offline Training}
During the training phase, we use each of the labeled training instances to construct a covariance matrix for each action. In order to make the covariance matrices more discriminative, the symmetric positive definite (SPD) matrix dimensionality reduction method~\cite{harandi2014manifold} is adopted to learn a projection matrix. The original covariance matrices are then projected onto a low dimensional but more discriminative space. As described in~\cite{harandi2014manifold}, the projection matrix learning can be expressed as the following minimization problem
\begin{equation}\label{LearningProjectionMat}
\begin{array}{l}
{P^ * } = \mathop {\arg \min }\limits_{P \in {\mathbb{R}^{n \times m}}} \sum\limits_{i,j} {{A_{ij}}} {\delta ^2}\left( {{P^T}{X_i}P,{P^T}{X_j}P} \right)\;\\
\;s.t.\;{P^T}P = {I_m}
\end{array}
\end{equation}
where $X_i$ and $X_j$ are any two different covariance matrices in the training set. $A_{ij}$ is a real symmetric affinity matrix which encodes the structure of the original data. $n$ is the original covariance matrix dimension and $m$ is the projected covariance matrix dimension. $I_m$ is a $m \times m$ identity matrix. $\delta$ is the Stein metric function~\cite{srapositive} or AIRM metric function~\cite{pennec2006riemannian}. In our experiment, we use the Stein metric for training and testing. For any two SPD covariance matrices $X$ and $Y$, their Stein metric is defined as
\begin{equation}\label{SteinMetric}
{\delta ^2}\left( {X,Y} \right) = \ln \det \left( {\frac{{X + Y}}{2}} \right) - \frac{1}{2}\ln \det \left( {XY} \right).
\end{equation}
There are two advantages in using the projection method. First, it tends to render the covariance matrices more discriminative. Second, it projects the high dimensional matrix into a low dimensional matrix, so the subsequent estimation of the distance between the two matrices is computationally efficient.

\begin{figure}
  \centering
  \includegraphics[width=0.5\textwidth]{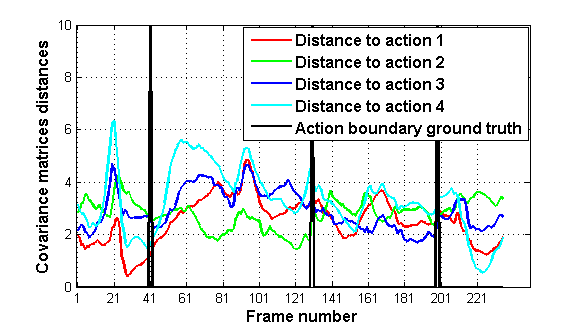}\\
  \caption{Minimum distances between online covariance matrices and training covariance matrices of each action class.}\label{DistanceCurve}
\end{figure}

\subsubsection{Online Recognition}
\begin{figure}
  \centering
  \includegraphics[width=0.5\textwidth]{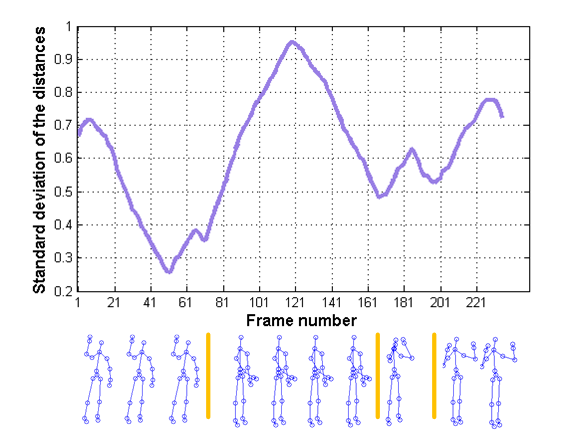}\\
  \caption{The standard deviation of the covariance distances.}\label{Std}
\end{figure}

During the online recognition phase, a continuous video stream consists of some actions with unknown order, and its labels are predicted frame by frame. At time $t$, we use the previous $t-1$ frames and current frame to construct a covariance matrix $C_t$. Then $C_t$ is projected to a low dimensional space using $P$ and  the distance between all the projected training covariance matrices and $C_t$ is compared. For each training class $l$, the $N_l$ distances are represented as $\left\{ {{d_{{C_t},{l_1}}}, \cdots ,{d_{{C_t},{l_{N_l}}}}} \right\}$. The final distance between $C_t$ and class $l$ is given by
\begin{equation}\label{MinDistance}
{d_{{C_t},l}} = \min \left\{ {{d_{{C_t},{l_1}}}, \cdots ,{d_{{C_t},{l_{N_l}}}}} \right\}.
\end{equation}
Thus, $L$ distances for each frame are generated as time progresses. At time $t$, the $L$ distances and their standard deviation are used to decide what action is being performed or determine whether there is a boundary between two consecutive actions. At the beginning, a covariance matrix using the first $t_0$ frames is initialized, and  an action label $l_{t_0}$ is given by the criterion
\begin{equation}\label{InitialDecision}
{l_{t_0}} = \arg \mathop {\min }\limits_l \left\{ {{d_{{C_{t_0}},l}}} \right\}_{l = 1}^L.
\end{equation}
From time $t_0+1$, if the standard deviation value of the $L$ distances is a local minima and a new action is detected by Eq.~\ref{InitialDecision}, action change is considered to take place. Otherwise, no action change happens. It is convenient to use the standard deviation of the $L$ distances to decide whether there exists a boundary between two actions. For instance, when the estimated distance at a given time is the minimum among others and the standard deviation is also large, this indicates that some specific action is being performed and there is no action change. In contrast, if a new action starts and the previous action ends, there exists a transitional stage, so all the estimated distances are similar and the standard deviation is relatively low. In~\cite{fanello2013keep} and~\cite{DeRosa2014}, a similar method is used on the SVM scores. Fig.~\ref{DistanceCurve} shows a segment of minimum distances between online covariance matrix and trained covariance matrices for each action class. Fig.~\ref{Std} shows the standard deviation of the distances in Fig.~\ref{DistanceCurve}. As can be seen, when it comes to an action change, the standard deviation value goes to a local minimal.

\subsection{Incremental leaning of weighted covariance matrices}
\label{IncrementalCovarianceLearning}
In previous methods, feature descriptors extracted from one video frame or video segments have been weighted equally. However, the contribution of each frame to action recognition varies as some frames are more discriminative than others. In this regard, the important frame should be weighted proportionally higher (frame-based weighting). Furthermore, during the online action recognition, recent frames nearer the current frame usually provide more information than past frames (temporal weighting). The proposed method incorporates these two weighting schemes in an efficient algorithm to update the covariance descriptor. Specifically, we assign two kinds of weights to each video frame - frame weight and temporal weight. There are a number of considerations in assigning the weights for each frame: 1) the frame weight should depend on the discriminative power of the frame with respect to the action and is independent of time; 2) the temporal weight of each frame should vary over time $t$ and recent frames should have higher time weights than past frames; 3) the weights should be computationally efficient and support incremental updating of the covariance matrix. We denote the frame weight of the $i_{th}$ frame as $\xi_i$. Similarly, at current time $t$, the temporal weight of the $i_{th}$ frame is denoted as $\omega _t^i$ is defined as follows:
\begin{equation}\label{TimeW}
\begin{array}{l}
\omega _t^i = g(t - i),\;\;\;i \in [1,t],\;\;\;\\
s.t.\;\\
g(0) = 1,\\
g(t + 1 - i) = \eta g(t - i),\; \eta \in [0,1].
\end{array}
\end{equation}
Then the final weight of the $i_{th}$ frame can be obtained as ${\psi _i} = {\xi _i}\omega _t^i$. With this weight setting, when a new frame at time $t+1$ is available, the covariance can be incrementally updated. Wu et al.~\cite{wu2012real} used a similar weighting scheme for tracking based on covariance descriptor, but only temporal weight was considered in their work. Next, we provide details of an efficient algorithm to update the weighted covariance matrix.

Let the feature vector extracted from the $i_{th}$ frame be ${f_i}$. The weighted feature vectors up to current time $t$ can be written as
\begin{equation}\label{FVectorSetTimeT}
{F_t} = {\left\{ {{f_i},\;{\psi _i}} \right\}_{i = 1, \cdots ,t}}
\end{equation}

Let $C_t$ and $\mu_t$ be the weighted covariance and the weighted mean of the feature vectors up to time $t$. The
formulation of $C_t$ and $\mu_t$ are as follows~\cite{price1972extension}:
\begin{equation}\label{CovT}
{C_t} = \frac{1}{{1 - \tilde \omega _t^2}}\sum {_{i= 1}^t\frac{\psi_i}{{{{\hat \omega }_t}}}({f_i} - {\mu _t}){{({f_i} - {\mu _t})}^T}},
\end{equation}
and
\begin{equation}\label{MeanT}
{\mu _t} = \frac{1}{{{{\hat \omega }_t}}}\sum {_{i = 1}^t} \psi_i{f_i},
\end{equation}
where
\begin{equation*}
{{\hat \omega }_t} = \sum {_{i = 1}^t} \psi_i \;\;\;\;\;\;\;\; \tilde \omega _t^2 = \sum {_{i = 1}^t} {\left( {\frac{{\psi_i}}{{{{\hat \omega }_t}}}} \right)^2}.
\end{equation*}

Our goal, therefore, is to efficiently compute the new covariance $C_{t+1}$ and mean $\mu_{t+1}$ using $C_t$ and $\mu_t$ when given $f_{t+1}$, ${\hat \omega }_{t}$, and $\tilde \omega _{t}^2$, without explicitly recomputing them from the data ${F_{t+1}}$. We give the incremental
covariance update rule in the following Theorem 1.
\newtheorem{CovUpdateRule}{Theorem}\label{CovUpdateRule}
\begin{CovUpdateRule}
Given $C_t$, $\mu_t$, ${\hat \omega }_{t}$, $\tilde \omega _{t}^2$, $f_{t+1}$,
$\xi_i$, $\omega _t^i = {g(t-i)}$, and $\eta$, the relation between
$C_t$ and $C_{t+1}$ can be written as
\begin{equation}\label{IncrementalCovRule}
\begin{array}{l}
{C_{t + 1}}{\rm{ = }}\frac{1}{{2\eta {{\hat \omega }_t}{\xi _{t + 1}} + {\eta ^2}{{\hat \omega }_t}^2\left( {1 - \tilde \omega _t^2} \right)}}\left\{ {\left[ {\eta {{\hat \omega }_t}\left( {1 - \tilde \omega _t^2} \right){C_t}} \right]\left( {{\xi _{t + 1}} + \eta {{\hat \omega }_t}} \right)} \right.\\
\;\;\;\;\;\;\;\;\; + \left. {\frac{{\eta {{\hat \omega }_t}\left( {{\xi _{t + 1}}^2 + \eta {{\hat \omega }_t}{\xi _{t + 1}}} \right)}}{{\eta {{\hat \omega }_t} + {\xi _{t + 1}}}}({f_{t + 1}} - {\mu _t}){{({f_{t + 1}} - {\mu _t})}^T}} \right\}
\end{array}
\end{equation}
and the relation between $\mu_t$ and $\mu_{t+1}$ is as follows
\begin{equation}\label{IncrementalMeanRule}
{\mu _{t{\rm{ + }}1}}{\rm{ = }}\frac{{\eta {{\hat \omega }_t}{\mu _t}{\rm{ + }}{\xi _{t + 1}}{f_{t + 1}}}}{{\eta {{\hat \omega }_t}{\rm{ + }}{\xi _{t + 1}}}},
\end{equation}
\end{CovUpdateRule}

In order to make the proof of Theorem 1 concise we first give some
lemmas. The proofs of all the lemmas appear in Appendix A.

\begin{lemma}
\label{LemmaSum}
If $\omega _t^i$ is given by Eq.~\ref{TimeW}, then we have
${{\hat \omega }_{t + 1}} = \eta {{\hat \omega }_t} + {\xi _{t + 1}}$ and $\tilde \omega _{t + 1}^2 = \frac{{{{\hat \omega }_t}^2{\eta ^2}\tilde \omega _t^2 + {\xi _{t + 1}}^2}}{{{{\left( {{\xi _{t + 1}} + \eta {{\hat \omega }_t}} \right)}^2}}}$.
\end{lemma}

\begin{lemma}
\label{Transpose}
$\sum {_{i = 1}^t{\xi _i}\omega _{t + 1}^i({f_t} - {\mu _t})}=0$,
and ${\sum {_{i = 1}^t{\xi _i}\omega _{t + 1}^i({f_i} - {\mu _t})} ^T} = 0$.
\end{lemma}
\begin{lemma}
\label{TransposeMul}
$\left( {{\mu _t} - {\mu _{t + 1}}} \right){\left( {{\mu _t} - {\mu _{t + 1}}} \right)^T} = \frac{{{\xi _{t + 1}}^2\left( {{\mu _t} - {f_{t + 1}}} \right)}}{{{{\left( {{\xi _{t + 1}} + \eta {{\hat \omega }_t}} \right)}^2}}}$
\end{lemma}
\begin{lemma}
\label{FeatureAndMuTransposeMul}
$\sum {_{i = 1}^t{\xi _i}\omega _{t + 1}^i({f_i} - {\mu _{t + 1}}){{({f_i} - {\mu _{t + 1}})}^T}}\\ = \eta \left( {1 - \tilde \omega _t^2} \right){{\hat \omega }_t}{C_t} + \frac{{\eta {{\hat \omega }_t}{\xi _{t + 1}}^2\left( {{\mu _t} - {f_{t + 1}}} \right){{\left( {{\mu _t} - {f_{t + 1}}} \right)}^T}}}{{{{\left( {{\xi _{t + 1}} + \eta {{\hat \omega }_t}} \right)}^2}}}$
\end{lemma}
Now, we give the proofs of Eq.~\ref{IncrementalCovRule} and Eq.~\ref{IncrementalMeanRule}.
According to the definition by Eq.~\ref{CovT} and Eq.~\ref{MeanT}, we have
\begin{equation}\label{CovTplusOne}
{C_{t + 1}} = \frac{1}{{1 - \tilde \omega _{t + 1}^2}}\sum {_{i = 1}^{t + 1}\frac{{{\xi _i}\omega _{t + 1}^i}}{{{{\hat \omega }_{t + 1}}}}({f_i} - {\mu _{t+ 1}}){{({f_i} - {\mu _{t + 1}})}^T}},
\end{equation}
and
\begin{align}
\label{MeanTplusOne}
 \mu_{t+1}=\dfrac{1}{\hat{\omega}_{t=1}}\sum_{i=1}^{t+1}\xi_i\omega_{t+1}^{i}f_i
\end{align}

\textbf{\emph{Proof of Eq.~\ref{IncrementalMeanRule}:}}
\begin{equation*}
\begin{array}{l}
{\mu _{t + 1}} = \frac{1}{{{{\hat \omega }_{t + 1}}}}\sum {_{i = 1}^{t + 1}} {\xi _i}\omega _{t + 1}^i{f_i}\\
\;\;\;\;\;\;\; = \frac{1}{{{{\hat \omega }_{t + 1}}}}\left\{ {\sum {_{i = 1}^t} {\xi _i}\omega _{t + 1}^i{f_i} + {\xi _{t + 1}}\omega _{t + 1}^{t + 1}{f_{t + 1}}} \right\}\\
\;\;\;\;\;\;\; = \frac{1}{{{{\hat \omega }_{t + 1}}}}\left\{ {\eta \sum {_{i = 1}^t} {\xi _i}\omega _t^i{f_i} + {\xi _{t + 1}}{f_{t + 1}}} \right\}\\
\;\;\;\;\;\;\; = \frac{1}{{{{\hat \omega }_{t + 1}}}}\left\{ {\eta {\mu
_t}{{\hat \omega }_t} + {\xi _{t + 1}}{f_{t + 1}}}
\right\}\footnote{1}\\
\;\;\;\;\;\;\; = \frac{{\eta {\mu _t}{{\hat \omega }_t} + {\xi _{t + 1}}{f_{t +
1}}}}{{{\xi _{t + 1}} + \eta {{\hat \omega
}_t}}}\footnote{2}
\end{array}
\footnotetext[1]{Using definition by Eq.\ref{MeanT}.}
\footnotetext[2]{Using Lemma.\ref{LemmaSum}.}
\end{equation*}
\textbf{\emph{Proof of Eq.~\ref{IncrementalCovRule}:}}
From Eq.~\ref{CovTplusOne}, we have
\begin{equation*}
\begin{array}{l}
{{\hat \omega }_{t + 1}} \cdot (1 - \tilde \omega _{t + 1}^2){C_{t + 1}} = \sum {_{i = 1}^{t + 1}{\xi _i}\omega _{t + 1}^i({f_i} - {\mu _{t + 1}}){{({f_i} - {\mu _{t + 1}})}^T}} \\
\;\;\;\;\;\;\;\;\;\;\;\;\;\;\;\;\;\;\;\;\;\;\;\;\;\;\; = \sum {_{i = 1}^t{\xi _i}\omega _{t + 1}^i({f_i} - {\mu _{t + 1}}){{({f_i} - {\mu _{t + 1}})}^T}} \\
\;\;\;\;\;\;\;\;\;\;\;\;\;\;\;\;\;\;\;\;\;\;\;\;\;\;\;\;\;\; + {\xi _{t + 1}}\omega _{t + 1}^{t + 1}({f_{t + 1}} - {\mu _{t + 1}})({f_{t + 1}} - {\mu _{t + 1}})^T\\
\;\;\;\;\;\;\;\;\;\;\;\;\;\;\;\;\;\;\;\;\;\;\;\;\;\;\; = \eta \left( {1 - \tilde \omega _t^2} \right){{\hat \omega }_t}{C_t}\\
 \;\;\;\;\;\;\;\;\;\;\;\;\;\;\;\;\;\;\;\;\;\;\;\;\;\;\;\;\;\;+ \frac{{\eta {{\hat \omega }_t}{\xi _{t + 1}}^2\left( {{\mu _t} - {f_{t + 1}}} \right){{\left( {{\mu _t} - {f_{t + 1}}} \right)}^T}}}{{{{\left( {{\xi _{t + 1}} + \eta {{\hat \omega }_t}} \right)}^2}}}\\
\;\;\;\;\;\;\;\;\;\;\;\;\;\;\;\;\;\;\;\;\;\;\;\;\;\;\;\;\;\; + \frac{{{\eta ^2}{{\hat \omega }_t}^2{\xi _{t + 1}}\left( {{\mu _t} - {f_{t + 1}}} \right){{\left( {{\mu _t} - {f_{t + 1}}} \right)}^T}}}{{{{\left( {{\xi _{t + 1}} + \eta {{\hat \omega }_t}} \right)}^2}}}\footnote{3}
\end{array}
\footnotetext[3]{Using Lemma.\ref{FeatureAndMuTransposeMul} and Eq.\ref{IncrementalMeanRule}.}
\end{equation*}
therefore,
\begin{equation*}
\begin{array}{l}
{C_{t + 1}}{\rm{ = }}\frac{1}{{2\eta {{\hat \omega }_t}{\xi _{t + 1}} + {\eta ^2}{{\hat \omega }_t}^2\left( {1 - \tilde \omega _t^2} \right)}}\left\{ {\left[ {\eta {{\hat \omega }_t}\left( {1 - \tilde \omega _t^2} \right){C_t}} \right]\left( {{\xi _{t + 1}} + \eta {{\hat \omega }_t}} \right)} \right.\\
\;\;\;\;\;\;\;\;\; + \left. {\frac{{\eta {{\hat \omega }_t}\left( {{\xi _{t + 1}}^2 + \eta {{\hat \omega }_t}{\xi _{t + 1}}} \right)}}{{\eta {{\hat \omega }_t} + {\xi _{t + 1}}}}({f_{t + 1}} - {\mu _t}){{({f_{t + 1}} - {\mu _t})}^T}} \right\}\footnote{4}
\end{array}
\footnotetext[4]{Using Lemma.\ref{LemmaSum}.}
\end{equation*}

In Eq.~\ref{IncrementalCovRule} and Eq.~\ref{IncrementalMeanRule}, if all the
frame weights
and temporal weights are equal to 1, we obtain
\begin{equation}\label{IncrementalCovRuleWeight1}
{C_{t + 1}} = \frac{{t - 1}}{t}{C_t} + \frac{{\left( {{f_{t + 1}} - {\mu _t}} \right){{\left( {{f_{t + 1}} - {\mu _t}} \right)}^T}}}{{t + 1}}
\end{equation}
and
\begin{equation}\label{IncrementalMeanRuleWeight1}
{\mu _{t + 1}} = \frac{{t{\mu _t} + {f_{t + 1}}}}{{t + 1}},
\end{equation}
which is the most common formulation.

\section{Experimental Results and Discussions}\label{experimental}
In this section, we give the experimental results on MSRC-12 Kinect Gesture dataset~\cite{fothergill2012instructing} and our newly collected online action recognition dataset. Skeleton was the only data used for our experiment and we adopt latency, miss rate and error rate to evaluate the performance. Similar to~\cite{ellis2013exploring} and~\cite{kviatkovsky2014online},  latency was used as one of the criteria to evaluate our algorithm. If the interval between the frame a subject begins the action and the frame our algorithm classifies the action is $h$, and the whole length of an action is $H$, then the latency of this action is defined as $\frac{h}{H}$. The miss rate and error rate were used to measure audio diarization error~\cite{tranter2006overview}. We argue that our online action recognition is largely similar to the audio diarization process, so we also uses these two criteria. If an action appears $n$ times in a video sequence, our algorithm can detect $m$ times, then the miss rate is defined as $\frac{{n - m}}{n}$. If an action can be detected and the length of this action is $W$, but during these $W$ frames, there are $w$ frames detected as other actions, then the error rate is defined as $\frac{w}{W}$. The latency was used to evaluate the sensitivity of our method while the miss rate and error rate are adopted to evaluate the accuracy of our method. 

For the purpose of demonstrating the efficacy of the temporal weight and frame weight, the results without the weights, represented as ``No $\omega^i$'' and ``No $\xi_i$'' are reported respectively in the following tables. 

The algorithms are compared with other two methods~\cite{kviatkovsky2014online,hussein2013human}, and both of them are based on covariance descriptors and sliding window. In~\cite{kviatkovsky2014online}, the online action recognition is accomplished by nearest neighbor search based on a sliding window. Here, we set the sliding window size to 40 frames (most of the actions in the testing datasets are accomplished by 35--50 frames). In order to compare with~\cite{hussein2013human}, their work is modified to make a prediction by SVM based on a 40-frame sliding window.

\subsection{Skeleton data normalization}
\label{SkeletonNormalization}
Generally, human actions usually involve body movement such as displacement and angle change, and different people have different body sizes. So it is important to normalize the skeleton data and make it body size and view angle invariant. In our work, the normalized relative 3D coordinates are used instead of the original absolute coordinates. In the original skeleton video data, each joint $i$ has three coordinates which can be represented as ${p_i}\left( t \right) = \left( {{x_i}\left( t\right),{y_i}\left( t \right),{z_i}\left( t \right)} \right)$ at frame $t$. Here, the \emph{hip center} is set as our new origin, and the new coordinates of other joints can be obtained from the difference between them and the hip center. In order to remove the body size variant, each new coordinates are normalized by the distance between the shoulder center joint and spline joint calculated using the original coordinates. Then the new coordinates of each joints $i$ can be expressed as 
${{p'}_i}\left( t \right) = \left(
{\frac{{{x_i}\left( t \right) - {x_{hip}}\left( t
\right)}}{d},\frac{{{y_i}\left( t \right) - {y_{hip}}\left( t
\right)}}{d},\frac{{{z_i}\left( t \right) - {z_{hip}}\left( t \right)}}{d}}
\right)$.

\subsection{Covariance descriptor}
\label{OurCovDes}
The update process could face numerical instability. Hence all the covariance matrices are forced to be positive definite because conventional covariance matrices distance metric methods such as Affine-Invariant distance metric~\cite{pennec2006riemannian}, Log-Euclidean metric~\cite{arsigny2007geometric} and S-Divergence root~\cite{srapositive} could be unstable. Here, the simplest skeleton 3D coordinates were adopted as our feature, which is a $(K-1) \times 3$ dimension vector. $K$ is the joints number of each frame and the coordinates of normalized origin is $(0,0,0)$, which was removed from our final feature representation. In order to avoid singularity of the covariance matrices, a small perturbation is added to the covariance matrices~\cite{wang2012covariance}.

\subsection{Temporal weight and frame weight}
\begin{figure}
  \centering
  \subfigure[Temporal weight varies with different time ($\eta=0.9$).]{
    \label{TimeWeight} 
    \includegraphics[width=0.45\textwidth]{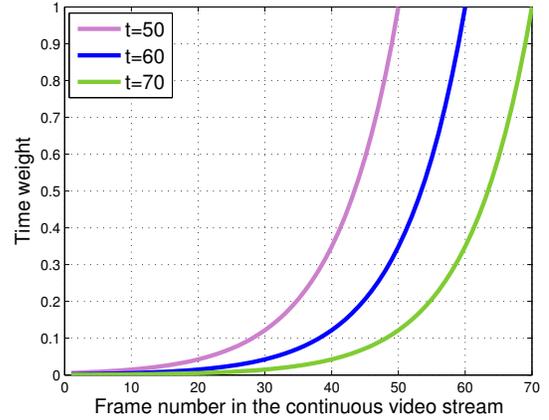}}
  \subfigure[Frame weight of different frames in an action instance.]{
    \label{FrameWeight} 
    \includegraphics[width=0.45\textwidth]{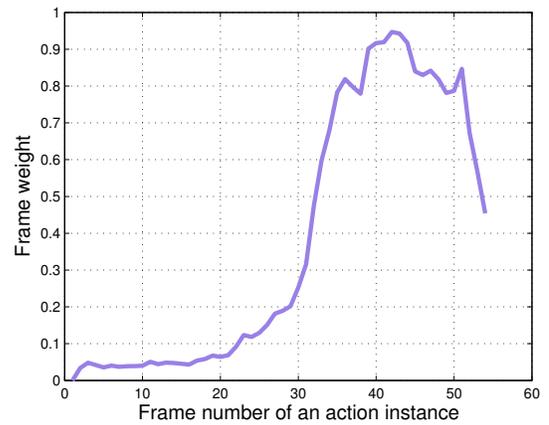}}
  \caption{Temporal weight and frame weight.}
  \label{Weight} 
\end{figure}

As illustrated in section~\ref{IncrementalCovarianceLearning},
the temporal weight of each frame should vary over time $t$;
and the frames from the current time should have higher temporal weights than
previous frames. Intuitively and based on these two points, the
power function is applied to compute our time weight. At time $t$, our temporal weight for
the $t_{th}$ frame is defined as
\begin{equation}\label{SpecificTimeW}
\omega _t^i = {\eta ^{t - i}},\;\;\;i \in [1,t],\;\eta  \in [0,1].
\end{equation}
We plot the temporal weight as it varies over time $t$ when
$\eta$ is set to 0.9 in Fig.~\ref{TimeWeight}. With the progression of time,
the temporal weight of past frames is effectively attenuated and
the most current frame is always assigned temporal weight of 1. Note that
other functions that satisfy the conditions in Eq.~\ref{TimeW} can be also used.

If a frame is important to a specific action, it is usually
quite different to the neutral pose (the position of joints where the bones that make up the joints are placed in the optimal position for maximal movement). Therefore, we use the average relative
distance of each joint between current frame and manually selected neutral pose to represent the frame weight. As all the actions on the two datasets have similar neutral pose, we only selected once for the neutral pose.
Fig.~\ref{FrameWeight} shows an example and it can be seen that the most initial
frames and the ending frames have relatively smaller frame weight because these
frames are similar to natural pose. They contribute little to recognize an
action.

\subsection{Experimental Results on MSRC-12 Kinect Gesture Dataset}

The MSRC-12 Gesture dataset~\cite{fothergill2012instructing}
contains 594 sequences, and there are more than 700000 frames
collected from 30 people performing 12 classes of gestures.
The frames were manually labeled into 6244 gesture instances. Twenty human body
skeleton joints ($K=20$) were captured by the Microsoft Kinect system. The body
poses were captured at a sample rate of 30Hz with an accuracy approximately two
centimeters in joint positions. The 12 actions are: lift outstretched arms, duck, push right, goggles, wind it up, shoot, bow, throw, had enough, change weapon, beat both.


In this dataset, the participants were provided with three
instruction modalities or their combination to perform
gestures in order to research various methods of teaching human on
how to perform different gestures. The three instruction modalities are i) text
descriptions, ii) image sequences, and iii) video demos. The two combination of
the three modalities are images with text and video with text. So the whole
dataset can be divided into 5 parts. For each part, the ``leave-person-out'' protocol
is used. Each time the training instances are used to calculate training
covariance matrices, and use these matrices to learn a projection matrix $P$ as
introduced in Section~\ref{Overview}. The training covariance matrices and
online updated covariance matrix are projected to a lower dimensional space by
the projection matrix $P$ before measuring their distances. In the testing
phase, $\eta$ is set to 0.95 to obtain the best result; the initial frame number
is 30. The action instances of the test subject are randomly stitched together
by each time. For each action in each modality, its average latency is obtained.
We finally average the 3 evaluation criteria on the 5 different modalities and the results
are shown in Table~\ref{LMRResultsMSRC-12Weight}. The results without temporal weight (NO $\omega^i$) or frame weight (NO $\xi_i$) are also presented. From Table~\ref{LMRResultsMSRC-12Weight}, it can be observed
that most action latency and miss rate are very low with our weighting function. This verifies the effectiveness of our temporal weight function. At a time, the current frame is assigned the maximum weight, when a new action is available, it can be quickly detected because the influence of past frames are diminished. The
comparison results between our methods and other methods are shown in Table~\ref{LMRResultsMSRC12Compare}, as can be seen, our method can perform better than other sliding window based methods.
\begin{table*}[!htb]
\caption{Average latency, miss rate and error rate with/without weighting of all the actions on MSRC-12 gesture dataset.}
\label{LMRResultsMSRC-12Weight}\centering
\begin{tabular}{|p{3.0cm}<{\centering}|p{0.6cm}<{\centering}|p{0.6cm}<{\centering}|p{1.0cm}<{\centering}|p{0.6cm}<{\centering}|p{0.6cm}<{\centering}|p{1.0cm}<{\centering}|p{0.6cm}<{\centering}|p{0.6cm}<{\centering}|p{1.0cm}<{\centering}|}
\hline
\multirow{2}{*}{Evaluation} & \multicolumn{3}{c|}{Latency (\%)} & \multicolumn{3}{c|}{Miss rate (\%)} & \multicolumn{3}{c|}{Error rate (\%)}\\
\cline{2-10}
& No $\omega^i$ & No $\xi_i$ & With $\omega^i$\&$\xi_i$ & No $\omega^i$ & No $\xi_i$ & With $\omega^i$\&$\xi_i$ & No $\omega^i$ & No $\xi_i$ & With $\omega^i$\&$\xi_i$ \\
\hline
\textbf{Average score} &  --   &  31.6    & \textbf{29.0 }& 92 &  13.8   &  \textbf{9.4} &  91.7    & 54.1 &     \textbf{51.6}\\
\hline   
\end{tabular}
\end{table*}

\begin{table*}[!htbp]
\caption{Comparison of our online action recognition results with~\cite{kviatkovsky2014online} and~\cite{hussein2013human} on  MSRC-12 gesture dataset.}
\label{LMRResultsMSRC12Compare}\centering
\begin{tabular}{|p{1.6cm}<{\centering}|p{0.5cm}<{\centering}|p{0.5cm}<{\centering}|p{0.6cm}<{\centering}|p{0.5cm}<{\centering}|p{0.5cm}<{\centering}|p{0.6cm}<{\centering}|p{0.5cm}<{\centering}|p{0.5cm}<{\centering}|p{0.6cm}<{\centering}|}
\hline
\multirow{3}{*}{Evaluation} & \multicolumn{3}{c|}{Latency (\%)} & \multicolumn{3}{c|}{Miss rate (\%)} & \multicolumn{3}{c|}{Error rate (\%)}\\
\cline{2-10}
& \cite{kviatkovsky2014online} & \cite{hussein2013human} & Ours & \cite{kviatkovsky2014online} & \cite{hussein2013human} & Ours & \cite{kviatkovsky2014online} & \cite{hussein2013human} & Ours \\
\hline
Average & 52 & 41.3 & \textbf{29.0}  & 20.7 & 15.2 &  \textbf{9.4} & 63.8 & 57.4 & \textbf{51.6}\\
\hline
\end{tabular}
\end{table*}



\subsection{Experimental Results on the Newly Collected Online Action Recognition Dataset}
As far as we know, there is nearly no essentially benchmark dataset for online action recognition. 
Thus we collected a dataset (Online Action3D\footnote{The dataset will be released with this paper.}) that human subjects perform actions continuously and naturally using the Microsoft Kinect V2.0~\texttrademark(25 human body skeleton joints are captured, $K=25$). The original actions of MSR Action3D dataset are used here. Firstly, 20 participants performed all the actions 5 or 6 times, and these samples are for training. Then, each participant performed the 20 actions continuously 1 or 2 times in random order, and these continuous action sequences are for online testing. The actions were captured at a sample rate of 20Hz. The ground truth of action segments were marked manually. Table~\ref{LMRResultsOnlineAction3D} gives the results with/without our weight function. As can be seen, experimental results on our new dataset also verify the efficacy of our method. Table~\ref{LMRResultsOnlineAction3DCompare} presents the comparison with other methods, it also shows the superiority of our method. Our method can handle the transition between any two different actions well.

\begin{table*}[!htb]
\caption{Average latency, miss rate and error rate with/without weighting of all the actions on Online Action3D dataset.}
\label{LMRResultsOnlineAction3D}\centering
\begin{tabular}{|p{3.0cm}<{\centering}|p{0.6cm}<{\centering}|p{0.6cm}<{\centering}|p{1.0cm}<{\centering}|p{0.6cm}<{\centering}|p{0.6cm}<{\centering}|p{1.0cm}<{\centering}|p{0.6cm}<{\centering}|p{0.6cm}<{\centering}|p{1.0cm}<{\centering}|}
\hline
\multirow{2}{*}{Evaluation} & \multicolumn{3}{c|}{Latency (\%)} & \multicolumn{3}{c|}{Miss rate (\%)} & \multicolumn{3}{c|}{Error rate (\%)}\\
\cline{2-10}
& No $\omega^i$ & No $\xi_i$ & With $\omega^i$\&$\xi_i$ & No $\omega^i$ & No $\xi_i$ & With $\omega^i$\&$\xi_i$ & No $\omega^i$ & No $\xi_i$ & With $\omega^i$\&$\xi_i$ \\
&&&&&&&&&\\
\hline
Average score & --   &  36.1  &  32.5 & 91.2  & 30.1  &  28.6  & 93.8  & 55.8  & 57.2 \\
\hline
\end{tabular}
\end{table*}


\begin{table*}[!htbp]
\caption{Comparison of our online action recognition results with~\cite{kviatkovsky2014online} and~\cite{hussein2013human} on  Online Action3D dataset.}
\label{LMRResultsOnlineAction3DCompare}\centering
\begin{tabular}{|p{1.6cm}<{\centering}|p{0.5cm}<{\centering}|p{0.5cm}<{\centering}|p{0.6cm}<{\centering}|p{0.5cm}<{\centering}|p{0.5cm}<{\centering}|p{0.6cm}<{\centering}|p{0.5cm}<{\centering}|p{0.5cm}<{\centering}|p{0.6cm}<{\centering}|}
\hline
\multirow{3}{*}{Evaluation} & \multicolumn{3}{c|}{Latency (\%)} & \multicolumn{3}{c|}{Miss rate (\%)} & \multicolumn{3}{c|}{Error rate (\%)}\\
\cline{2-10}
& \cite{kviatkovsky2014online} & \cite{hussein2013human} & Ours & \cite{kviatkovsky2014online} & \cite{hussein2013human} & Ours & \cite{kviatkovsky2014online} & \cite{hussein2013human} & Ours \\
\hline
Average & 40.6 & 34.7 & \textbf{32.5}  & 49.5 & 42.3 &  \textbf{28.6} & 67.4 & 61.8 & \textbf{57.2}\\
\hline
\end{tabular}
\end{table*}

\subsection{Efficiency of the Weighted Covariance Learning}
\begin{figure}
  \centering
  \includegraphics[width=0.5\textwidth]{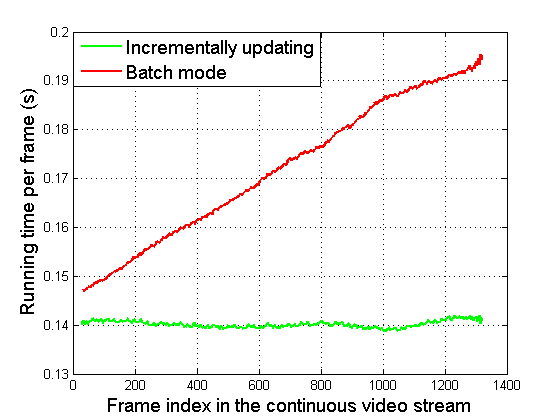}\\
  \caption{Running time of our online updating mode and batch mode for each frame.}\label{RunTime}
\end{figure}

Note that our method only uses skeleton data and operates in near real-time. Fig.~\ref{RunTime} shows the running time of each frame (initial frame number is 30). As can be seen, if all the covariance matrices are computed in batch mode, the running time will grow rapidly with video frames. When it comes to a very long video, processing one frame will take too much time. In contrast, our online incrementally updating manner can maintain a constant time for each frame, no matter how long the video. 

\section{Conclusion}\label{conclusion}
In this paper, we presented a fast online action recognition method based on incremental learning of weighted covariance descriptors. The covariance descriptor is robust to noise and can be updated efficiently using frame based features. We used a temporal weight function to effectively accumulate information with the evolution of time. In our current work, we adopted the average relative distance of each joint between current frame and neutral pose to represent frame weight, which suppresses the effect of the neutral poses and improve the recognition performance. In this paper, the neutral pose is selected manually, how to develop a neutral pose model which can recover the corresponding neutral pose from any frame is one of the future works. Another issue is that when some high dimensional feature vectors are used, the dimension of covariance matrices is very high, leading to the distance between covariance matrices not stable, and this fact may affect the final recognition accuracy. How to get an accurate metric (such as kernel methods) between two covariance matrices is also our future task. 


\appendix

\section{Proof of the Lemmas}
\subsection{Proof of Lemma~\ref{LemmaSum}:}
\begin{equation*}
\begin{array}{l}
{{\hat \omega }_{t + 1}} = \sum {_{i = 1}^{t + 1}} {\xi _i}\omega _{t + 1}^i = \sum {_{i = 1}^t} {\xi _i}\omega _{t + 1}^i + {\xi _{t + 1}}\omega _{t + 1}^{t + 1}\\
\;\;\;\;\;\;\; = \eta \sum {_{i = 1}^t} {\xi _i}\omega _t^i + {\xi _{t + 1}}\\
\;\;\;\;\;\;\; = \eta {{\hat \omega }_t} + {\xi _{t + 1}}
\end{array}
\end{equation*}
and
\begin{equation*}
\begin{array}{l}
\tilde \omega _{t + 1}^2 = \sum {_{i = 1}^{t + 1}} {\left( {\frac{{{\xi _i}\omega _{t + 1}^i}}{{{{\hat \omega }_{t + 1}}}}} \right)^2} = \sum {_{i = 1}^t} {\left( {\frac{{{\xi _i}\omega _{t + 1}^i}}{{{{\hat \omega }_{t + 1}}}}} \right)^2} + {\left( {\frac{{{\xi _{t + 1}}}}{{{{\hat \omega }_{t + 1}}}}} \right)^2}\\
\;\;\;\;\;\;\; = \frac{1}{{{{\hat \omega }_{t + 1}}^2}}\left\{ {\sum {_{i = 1}^t} {{\left( {{\xi _i}\omega _{t + 1}^i} \right)}^2} + {\xi _{t + 1}}^2} \right\}\\
\;\;\;\;\;\;\; = \frac{1}{{{{\hat \omega }_{t + 1}}^2}}\left\{ {{\eta ^2}\sum {_{i = 1}^t} {{\left( {{\xi _i}\omega _t^i} \right)}^2} + {\xi _{t + 1}}^2} \right\}\\
\;\;\;\;\;\;\; = \frac{1}{{{{\hat \omega }_{t + 1}}^2}}\left\{ {{\eta ^2}{{\hat \omega }_t}^2\sum {_{i = 1}^t} {{\left( {\frac{{{\xi _i}\omega _t^i}}{{{{\hat \omega }_t}}}} \right)}^2} + {\xi _{t + 1}}^2} \right\}\\
\;\;\;\;\;\;\; = \frac{1}{{{{\hat \omega }_{t + 1}}^2}}\left\{ {{\eta ^2}{{\hat \omega }_t}^2\tilde \omega _t^2 + {\xi _{t + 1}}^2} \right\}\\
\;\;\;\;\;\;\; = \frac{{{\eta ^2}{{\hat \omega }_t}^2\tilde \omega _t^2 + {\xi _{t + 1}}^2}}{{{{\left( {\eta {{\hat \omega }_t} + {\xi _{t + 1}}} \right)}^2}}}
\end{array}
\end{equation*}
\subsection{Proof of Lemma~\ref{Transpose}:}
\begin{equation*}
\begin{array}{l}
\sum {_{i = 1}^t{\xi _i}\omega _{t + 1}^i({f_i} - {\mu _t})}  = \sum {_{i = 1}^t{\xi _i}\omega _{t + 1}^i{f_i}}  - \sum {_{i = 1}^t{\xi _i}\omega _{t + 1}^i{\mu _t}} \\
\;\;\;\;\;\;\;\;\;\;\;\;\;\;\;\;\;\;\;\;\;\;\;\;\;\;\;\;\;\; = \eta \sum {_{i = 1}^t{\xi _i}\omega _t^i{f_i} - \eta \sum {_{i = 1}^t{\xi _i}\omega _t^i{\mu _t}} } \\
\;\;\;\;\;\;\;\;\;\;\;\;\;\;\;\;\;\;\;\;\;\;\;\;\;\;\;\;\;\; = \eta {\mu _t}{{\hat \omega }_t} - \eta {\mu _t}{{\hat \omega }_t}\\
\;\;\;\;\;\;\;\;\;\;\;\;\;\;\;\;\;\;\;\;\;\;\;\;\;\;\;\;\;\; = 0,
\end{array}
\end{equation*}
and
\begin{equation*}
{\sum {_{i = 1}^t{\xi _i}\omega _{t + 1}^i({f_i} - {\mu _t})} ^T} = {\left\{ {\sum {_{i = 1}^t{\xi _i}\omega _{t + 1}^i({f_i} - {\mu _t})} } \right\}^T} = 0.
\end{equation*}
\subsection{Proof of Lemma~\ref{TransposeMul}:}
\begin{equation*}
\begin{array}{l}
{\mu _t} - {\mu _{t + 1}} = {\mu _t} - \frac{{\eta {{\hat \omega }_t}{\mu
_t}{\rm{ + }}{\xi _{t + 1}}{f_{t + 1}}}}{{\eta {{\hat \omega }_t}{\rm{ + }}{\xi
_{t + 1}}}}\footnote{6}\\
\;\;\;\;\;\;\;\;\;\;\;\;\;\; = \frac{{{\xi _{t + 1}}\left( {{\mu _t} - {f_{t + 1}}} \right)}}{{\eta {{\hat \omega }_t}{\rm{ + }}{\xi _{t+ 1}}}}.
\end{array}
\footnotetext[6]{Using definition by Eq.~\ref{IncrementalMeanRule}.}
\end{equation*}
\subsection{Proof of Lemma~\ref{FeatureAndMuTransposeMul}:}
\begin{equation*}
\begin{array}{l}
\sum {_{i = 1}^t{\xi _i}\omega _{t + 1}^i({f_i} - {\mu _{t + 1}}){{({f_i} - {\mu _{t + 1}})}^T}}\\
= \sum {_{i = 1}^t{\xi _i}\omega _{t + 1}^i({f_i} - {\mu _t} + {\mu _t} - {\mu _{t + 1}}){{({f_i} - {\mu _t} + {\mu _t} - {\mu _{t + 1}})}^T}} \\
= \sum {_{i = 1}^t{\xi _i}\omega _{t + 1}^i({f_i} - {\mu _t}){{({f_i} - {\mu _t})}^T}} \\
\;\;\;\; + \left\{ {\sum {_{i = 1}^t{\xi _i}\omega _{t + 1}^i({f_i} - {\mu _t})} } \right\}{({\mu _t} - {\mu _{t + 1}})^T}\\
\;\;\;\;+ \left\{ {\sum {_{i = 1}^t{\xi _i}\omega _{t + 1}^i{{({f_i} - {\mu _t})}^T}} } \right\}({\mu _t} - {\mu _{t + 1}})\\
\;\;\;\; + \left\{ {({\mu _t} - {\mu _{t + 1}}){{({\mu _t} - {\mu _{t + 1}})}^T}} \right\}\sum {_{i = 1}^t{\xi _i}\omega _{t + 1}^i} \\
= \eta \sum {_{i = 1}^t{\xi _i}\omega _t^i({f_i} - {\mu _t}){{({f_i} - {\mu _t})}^T}} \;\;\;\\
\;\;\;\; + \frac{{\eta {{\hat \omega }_t}{\xi _{t + 1}}^2\left( {{\mu _t} -
{f_{t + 1}}} \right){{\left( {{\mu _t} - {f_{t + 1}}} \right)}^T}}}{{{{\left(
{\eta {{\hat \omega }_t}{\rm{ + }}{\xi _{t + 1}}}
\right)}^2}}}\footnote{7}\\
= \eta \left( {1 - \tilde \omega _t^2} \right){{\hat \omega }_t}{C_t} +
\frac{{\eta {{\hat \omega }_t}{\xi _{t + 1}}^2\left( {{\mu _t} - {f_{t + 1}}}
\right){{\left( {{\mu _t} - {f_{t + 1}}} \right)}^T}}}{{{{\left( {\eta {{\hat
\omega }_t}{\rm{ + }}{\xi _{t + 1}}} \right)}^2}}}\footnote{8}.
\end{array}
\footnotetext[7]{Using Lemma~\ref{Transpose} and Lemma~\ref{TransposeMul}.}
\footnotetext[8]{Using Eq.~\ref{CovT}.}
\end{equation*}

%


\end{document}